\documentclass[conference]{IEEEtran}

\usepackage{graphicx}
\usepackage{float}
\usepackage{makecell}  

\usepackage{amsmath}
\usepackage{algorithm}
\usepackage[noend]{algpseudocode}
\usepackage{algpseudocode}
\usepackage{algorithm,algpseudocode,float}
\usepackage{lipsum}

\begin{document}

\title{Modified Self-Organized Task Allocation in a Group of Robots}

\author{\IEEEauthorblockN{Chang Liu}
\IEEEauthorblockA{McCormick School of Engineering\\
Mechanical Engineering\\
Northwestern University\\
Evanston, Illinois, 60208\\
Email: ChangLiu2016@u.northwestern.edu}}

\maketitle

\begin{abstract}

This paper introduces a modified self-organized task allocation algorithm, where robots are assigned to pick up one of the two types of object. This paper also demonstrates both algorithms by showing the simulation results of the conventional self-organized task allocation algorithm and the simulation results of its modification. 

\end{abstract}

\IEEEpeerreviewmaketitle


\bigskip
\section{\textbf{Introduction}}
  
  The conventional task allocation algorithm does not distinct the object types, which means there is only one type of object in the experiments. Robots are finally separated into two different groups: (1) foragers: robots that have more tendencies to retrieve objects; (2) loafers: robots that have more tendencies to stay in the central nest.
  
  In the modified task allocation algorithm, there exists multiple types of robots (in the experiments shown in this paper, there are two types of object). Robots leave the central nest and are assigned to pick up one certain type of object. Through the experiment, each robot will grow tendency to pick up only one out of the two types of object. Therefore, ideally, robots should be separated into three groups eventually: (1) foragers for object type 1: robots that have more tendencies to leave the nest and pick up objects of type 1; (2) foragers for object type 2: robots that have more tendencies to leave the nest and pick up objects of type 2; and (3) loafers: robots that have more tendency to stay in the nest.

	This paper is designed by the following sequence: Section II mainly discusses the movement and collision avoidance strategies of the robots; Section III discusses two sets of experiment parameters and environments; Section IV reviews the conventional task allocation algorithm and introduces the modified algorithm; Section V shows the simulation results and Section VI gives the conclusion.


\bigskip
\section{\textbf{Details About Robots Movement}}

The robots are always in one of the three phases: (1) searching phase; (2) returning phase; and (3) stopping phase. Each robot has its own timer and once the robot starts moving, its timer will start counting. If the robot can not find any object when its searching time is out, it will go back home itself. The corresponding movement and collision avoidance strategies during searching phase and returning phase will be introduced in the following subsections. When the robot returns home, it will be set to stopping phase, where robot keep checking whether it is able to leave the nest by comparing a randomly generated value and the leave nest probability $P_1$, which will be introduced in later section.

\subsection{\textbf{Movement And Collision Avoidance Strategies In Searching Phase}}

	Robots are initially placed inside the central nest. Once the task allocation algorithm starts, robots start to leave the central nest and start searching. 

\subsubsection{No Collision} 
	
	As the robots moving inside the experimental arena, if a robot does not collide with any other robots or any object, the robot will move forward. The heading direction will be slightly changed by an random error.
	
\subsubsection{Collide With Other Robots}

	If a robot is too close to another robot during searching phase, both robots will be bounced away from each other. For bouncing, robots will keep updating their heading directions to random values until they are not too close to each other. 
	
\subsubsection{Collide With Objects}

	Each robot has its own probabilities to pick up different types of object due to mechanical differences. If a robot is too close to an object, the robot will generate a random number between 0 and 1 and compare it with the certain predefined probability. If the robot is able to pick up the object, it will carry the object back home, which is the situation in the next subsection. If a robot is not able to pick up the object, it will be bounced away to another direction and keep searching new objects. 
	
\subsubsection{Collide With The Boundary Of The Central Nest}

	Since robots will search within the entire experimental arena, it is possible that a robot will try to pass the central nest zone. In this case, if the robot is carrying an object, it will be allowed to pass the boundary of the central nest and drop the object. However, if the robot is not carrying an object, it will be bounced away when it is too close to the boundary of the central nest. Bouncing will be the same as the one in subsection 2).
	
\subsubsection{Collide With The Boundary Of The Experimental Arena}
	
	If a robot is too close to the boundary of the experimental arena, it will be bounced back, whether it is carrying an object or not. Technically, if a robot is carrying an object, it will move towards the central nest. Bouncing will be the same as the one in subsection 2).

\subsection{\textbf{Movement And Collision Avoidance Strategies In Returning Phase}}

A robot will be in returning phase if it successfully picks up an object or its searching time is out. In both cases, the collision avoidance strategies are the same.

\subsubsection{No Collision} 
	
	If a robot is carrying the chosen object and does not collide with any other robots or any other objects, the robot will move towards the origin with the chosen object. If a robot is not carrying any object and does not collide with any other robots or any object, it will also move towards the origin and finally pass the boundary of the central nest.
	
\subsubsection{Collide With Other Robots}
	
	If a robot is too close to another robot, whether they are carrying objects or not, the robot will be bounced to the opposed direction.

\subsubsection{Collide With Objects}

	If a robot is too close to an object when returning to the nest, it will not pick up the object since the searching time is out for this searching iteration. Instead, the robot will move along the edge of the object to pass it. 
	
\subsubsection{Collide With The Boundary Of The Central Nest}

	When robots are in returning phase, they are free to pass the boundary of the central nest.

\subsubsection{Collide With The Boundary Of The Experimental Arena}

	It is unlikely to collide with the boundary of the experimental arena during returning phase, since robots will move towards the origin during returning phase.


\bigskip
\section{\textbf{Experiments}}

Two sets of experiment have been conducted, one for the original algorithm and one for the modified algorithm. Twenty experiments have been conducted for each set of experiment and results are combined. Experiment parameters are summarized in Section A and the experiment environment is shown in Section B. 

\subsection{\textbf{Experiment Parameters}}

\subsubsection{Experiment Set I}

The experiment parameters used for testing the original organism are summarized in Table \ref{table:exp_parameters_1_1} - Table \ref{table:exp_parameters_1_4}.
	
	Table \ref{table:exp_parameters_1_1} tells the group size of the robots and the group size of the objects. Once an object is brought home, another object of the same type will be replaced in the experimental arena at a random location. In this way, the total numbers of objects in the experiment arena for each type are constant.

		\begin{table}[H]
				\caption{Number of Robots and Objects}
				\centering
			\begin{tabular}{  c | c   }
				\Xhline{4\arrayrulewidth}   
					Type & Total Number   \\ 
				\hline
					Robots (Purple Robots)  & 15 \\ 
					Objects of Type 1 (Green Objects)  &  30 \\  
					Objects of Type 2 (Red Objects)  &  35 \\ 
				\Xhline{4\arrayrulewidth}   
			\end{tabular}
			\label{table:exp_parameters_1_1}  
		\end{table}
		
	Each robot is mechanically different from other robots so that each robot has different probabilities of picking up each type of object, which is indicated in Table \ref{table:exp_parameters_1_2}.
		
		\begin{table}[H]
				\caption{Probabilities Of Picking Up Each Type Of Object}
				\centering
			\begin{tabular}{  c | c   }
				\Xhline{4\arrayrulewidth}   
					Type Of Object To Be Picked Up  &  Probability Range   \\ 
				\hline
					Objects of Type 1 (Green Objects)  &  [0 1] \\  
					Objects of Type 2 (Red Objects)  &  [0 1] \\ 
				\Xhline{4\arrayrulewidth}   
			\end{tabular}
			\label{table:exp_parameters_1_2}  
		\end{table}
		
	Experiment timeout and robot's searching time are shown in Table \ref{table:exp_parameters_1_3}. Note that each robot will use its own timer during searching. Once a robot can not find any object after the searching time interval, this robot will return home, without influencing other robots. 
		
		\begin{table}[H]
				\caption{Time Interval}
				\centering
			\begin{tabular}{  c | c   }
				\Xhline{4\arrayrulewidth}   
					 Type &  Length   \\ 
				\hline
					Experiment Time Interval (Seconds)  &  180 \\  
					Searching Time for Each Robot (Seconds)  &  15 \\ 
				\Xhline{4\arrayrulewidth}   
			\end{tabular}
			\label{table:exp_parameters_1_3}  
		\end{table}

		The probability parameters of leaving the nest are summarized in Table \ref{table:exp_parameters_1_4}.
		
		\begin{table}[H]
				\caption{Probability Parameters Of Leaving The Nest}
				\centering
			\begin{tabular}{  c | c   }
				\Xhline{4\arrayrulewidth}   
					 Type &  Probability   \\ 
				\hline
					$P_{1-max}$  &  0.08 \\  
					$P_{1-min}$  &  0.002 \\
					$P_{1-initial}$  & 0.04 \\
					$\Delta$  &  0.0003  \\
				\Xhline{4\arrayrulewidth}   
			\end{tabular}
			\label{table:exp_parameters_1_4}  
		\end{table}

\subsubsection{Experiment Set II}

	The experiment parameters for testing the modified algorithm are summarized in Table \ref{table:exp_parameters_2_1} - Table \ref{table:exp_parameters_2_5}.
	
	Table \ref{table:exp_parameters_2_1} tells the group size of the robots and the group size of the objects. 

		\begin{table}[H]
				\caption{Number of Robots and Objects}
				\centering
			\begin{tabular}{  c | c   }
				\Xhline{4\arrayrulewidth}   
					Type & Total Number   \\ 
				\hline
					Robots (Purple Robots)  & 15 \\ 
					Objects of Type 1 (Green Objects)  &  30 \\  
					Objects of Type 2 (Red Objects)  &  35 \\ 
				\Xhline{4\arrayrulewidth}   
			\end{tabular}
			\label{table:exp_parameters_2_1}  
		\end{table}
		
	Each robot is mechanically different from other robots so that each robot has different probabilities of picking up each type of object, which is indicated in Table \ref{table:exp_parameters_2_2}.
		
		\begin{table}[H]
				\caption{Probabilities Of Picking Up Each Type Of Object}
				\centering
			\begin{tabular}{  c | c   }
				\Xhline{4\arrayrulewidth}   
					Type Of Object To Be Picked Up  &  Probability Range   \\ 
				\hline
					Objects of Type 1 (Green Objects)  &  [0 1] \\  
					Objects of Type 2 (Red Objects)  &  [0 1] \\ 
				\Xhline{4\arrayrulewidth}   
			\end{tabular}
			\label{table:exp_parameters_2_2}  
		\end{table}
		
	Experiment timeout and robot's searching time are shown in Table \ref{table:exp_parameters_2_3}. Note that each robot will use its own timer during searching. Once a robot can not find any object after the searching time interval, this robot will return home, without influencing other robots. 
		
		\begin{table}[H]
				\caption{Time Interval}
				\centering
			\begin{tabular}{  c | c   }
				\Xhline{4\arrayrulewidth}   
					 Type &  Length   \\ 
				\hline
					Experiment Time Interval (Seconds)  &  300 \\  
					Searching Time for Each Robot (Seconds)  &  25 \\ 
				\Xhline{4\arrayrulewidth}   
			\end{tabular}
			\label{table:exp_parameters_2_3}  
		\end{table}
		
	The probability parameters of leaving the nest and picking up each type of object are summarized in Table \ref{table:exp_parameters_2_4} and Table \ref{table:exp_parameters_2_5}. For the experiments shown in this paper, the probability Parameters of picking up each type of object are set to be the same.
	
		\begin{table}[H]
				\caption{Probability Parameters Of Leaving The Nest}
				\centering
			\begin{tabular}{  c | c   }
				\Xhline{4\arrayrulewidth}   
					 Type &  Probability   \\ 
				\hline
					$P_{1-max}$  &  0.08 \\  
					$P_{1-min}$  &  0.002 \\
					$P_{1-initial}$  & 0.04 \\
					$\Delta$  &  0.0015  \\
				\Xhline{4\arrayrulewidth}   
			\end{tabular}
			\label{table:exp_parameters_2_4}  
		\end{table}
		
		\begin{table}[H]
				\caption{Probability Parameters Of Picking Up Objects}
				\centering
			\begin{tabular}{  c | c   }
				\Xhline{4\arrayrulewidth}   
					 Type &  Probability   \\ 
				\hline
					$P_{obj-1-max}$,  $P_{obj-2-max}$ &  0.15 \\  
					$P_{obj-1-min}$,  $P_{obj-2-min}$ &  0.002 \\
					$P_{obj-1-initial}$,  $P_{obj-2-initial}$ & 0.075 \\
					$\Delta_{obj-1}$, $\Delta_{obj-2}$   &  0.0025  \\
				\Xhline{4\arrayrulewidth}   
			\end{tabular}
			\label{table:exp_parameters_2_5}  
		\end{table}

 
\subsection{\textbf{Experiment Environments}}
	
\begin{figure}[H]
			\centering
			\includegraphics[scale=0.36]{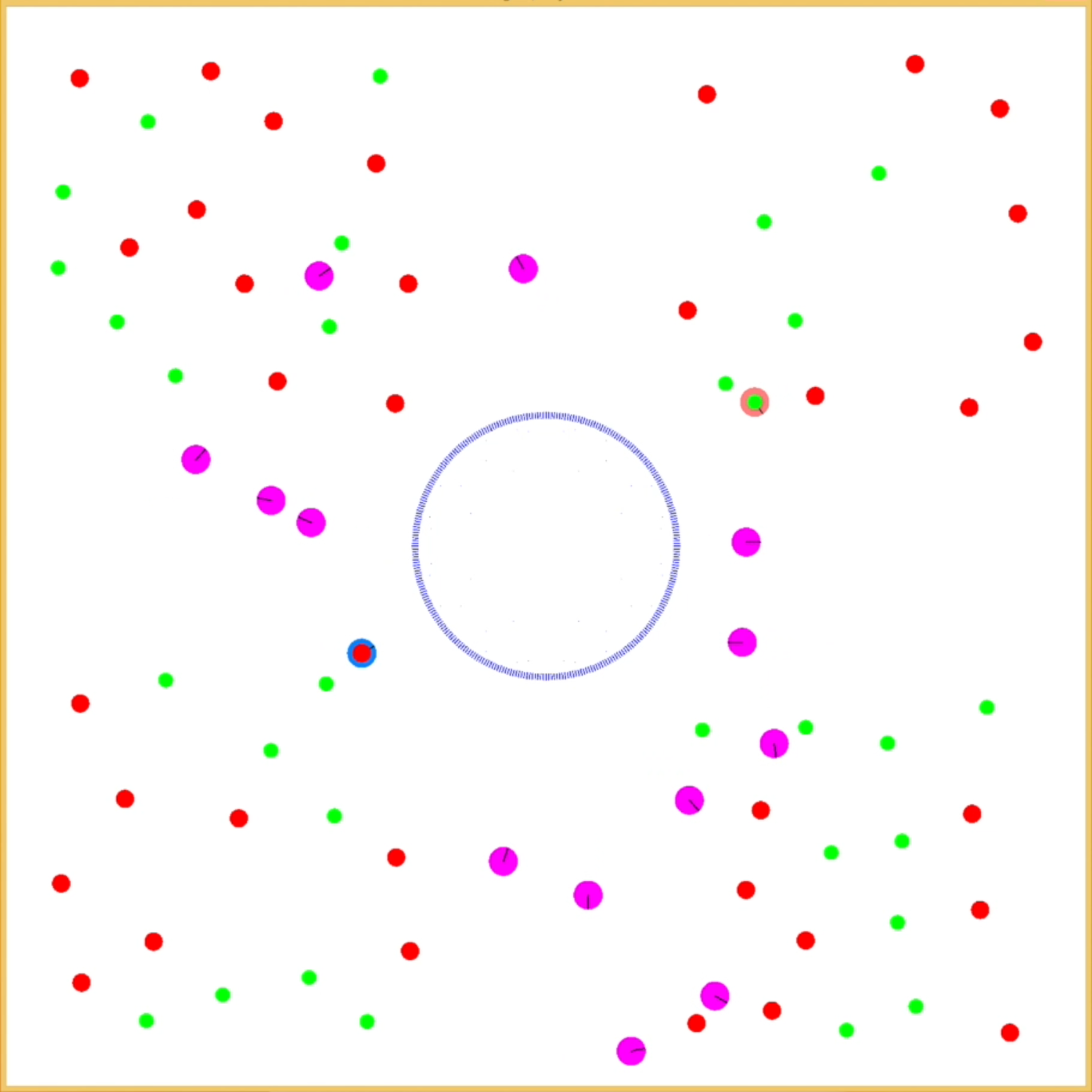}
			\caption{Experiment Environment}
			\label{fig:exp_1_envir}	
		\end{figure}

	The simulation screen shots are shown in Figure \ref{fig:exp_1_envir}.
	
	In each experiment, there is an nest located at the center of the experimental arena. Objects are randomly placed in the experimental arena. Objects do not collide with each other and there is no object placed inside the central nest. Robots are initially located inside the central nest and they use a random walk initially. When the experiment starts, robots will leave the nest and start searching objects in the experimental arena. Unless the robot successfully picks up an object or its searching time is out, no robot is allowed to return to the central nest. Once a robot picks up an object of type 1 (green object), it will turn orange from purple, and once a robot picks up an object of type 2 (red object), it will turn blue from purple. Once the robot returns home, it will be reset to purple. For each object that is brought home, another object of the same type will be placed at a random location in the experimental nest so that the total numbers of objects of each type in the experimental arena are constant through the experiment.

\bigskip
\section{\textbf{Algorithms}}

\subsection{\textbf{Previous Work}}

The conventional task-allocation algorithm treats all objects equally without telling the differences. The update of the probability of leaving the central nest, $P_1$, is done as shown in Algorithm 1, named Variable Delta Rule (VDR). The algorithm increments or decrements $P_1$ by a constant $\Delta$ multiplied by the number of consecutive successes or failures of leaving the central nest. It then bounds $P_1$ in the range [$P_{1-min}, P_{1-max}$].

\begin{algorithm}

	\caption{Variable Delta Rule (VDR). $P_1$ is the probability of leaving the nest, $\#_{succ}$ and $\#_{fail}$ are the number of consecutive successes and failures of leaving the nest.}
	\label{euclid}
	
	\begin{algorithmic}
	\State \textbf{initialization:} $\#_{succ} \gets 0$; $\#_{fail} \gets 0$; $P_1 \gets P_{1-initial}$
	\If {$success$} 
		 
		 $\#_{succ} \gets \#_{succ} + 1$
		 
		 $\#_{fail} \gets 0$
		 
		 $P_1 \gets min\{P_{1-max}, P_1 + \#_{succ} \cdot \Delta \} $
		 
	\EndIf
	
	\State \textbf{end if}

 	\If {$failure$} 
 	
		 $\#_{fail} \gets \#_{fail} + 1$
		 
		 $\#_{succ} \gets 0$
		 
		 $P_1 \gets max\{P_{1-min}, P_1 - \#_{fail} \cdot \Delta \} $ 
		 
	\EndIf
	
	\State \textbf{end if}
	
	\end{algorithmic}
\end{algorithm}

\subsection{\textbf{Modified Algorithm}}

	In the modified algorithm, robots will grow tendencies of picking up only one of the two types of object. The update of the probability of leaving the central nest, $P_1$, is done as shown in Algorithm 1, named Variable Delta Rule (VDR). The update of the probabilities of picking up certain type of object, $P_{obj-1}$ and $P_{obj-2}$, are done as shown in Algorithm 2. When a robot leaves home, it will be assigned to pick up one type of object, based on $P_{obj-1}$ and $P_{obj-2}$. The algorithm increments or decrements $P_{obj-1}$ and $P_{obj-2}$ by constant values $\Delta_{obj-1}$ and $\Delta_{obj-2}$ multiplied by the number of consecutive successes or failures of picking up the certain type of object. It then bounds $P_{obj-1}$ in the range [$P_{obj-1-min}, P_{obj-1-max}$] and bounds $P_{obj-2}$ in the range [$P_{obj-2-min}, P_{obj-2-max}$]

\begin{algorithm}

	\caption{$P_1$ is the probability of leaving the nest, $\#_{succ}$ and $\#_{fail}$ are the numbers of consecutive successes and failures of leaving the nest. $P_{obj-1}$ is the probability of picking up objects of type 1, $\#_{succ-obj-1}$ and $\#_{fail-obj-1}$ are the numbers of consecutive successes and failures of picking up objects of type 1. $P_{obj-2}$ is the probability of picking up objects of type 2, $\#_{succ-obj-2}$ and $\#_{fail-obj-2}$ are the numbers of consecutive successes and failures of picking up objects of type 2.}
	\label{euclid}
	
	\begin{algorithmic}
	\State \textbf{initialization:} $\#_{succ} \gets 0$; $\#_{fail} \gets 0$; $\#_{succ-obj-1} \gets 0$; $\#_{fail-obj-1} \gets 0$; $\#_{succ-obj-1} \gets 0$; $\#_{fail-obj-2} \gets 0$; $P_1 \gets P_{1-initial}$; $P_{obj-1} \gets P_{obj-1-initial}$; $P_{obj-2} \gets P_{obj-2-initial}$

	\If {$leave \ nest \ success$} 
		 
		 $\#_{succ} \gets \#_{succ} + 1$
		 
		 $\#_{fail} \gets 0$
		 
		 \If {$assigned \ to \ retrieve \ objects \ of \ type \ 1$}
		 		
		 		\If {$pick \ up \ success$}
		 		
		 				$\ \ \ \ \ \ \ \ \#_{succ-obj-1} \gets \#_{succ-obj-1} + 1$
		 				
		 				$\ \ \ \ \ \ \ \ \#_{fail-obj-1} \gets 0$
		 				
		 				$\ \ \ \ \ \ \ \ P_{obj-1} \gets min\{P_{obj-1-max}, $
		 				
		 				$\ \ \ \ \ \ \ \ \ \ \ \ \ \ \ \ \ \ \ \ \ \ \ \ P_{obj-1} + \#_{succ-obj-1} \cdot \Delta_{obj-1} \}$
		 		
		 		\EndIf
		 		\State \textbf{end if}
		 		
		 		\If {$pick \ up \ failure$}
		 					
		 				$\ \ \ \ \ \ \ \ \#_{succ-obj-1} \gets 0$
		 		
		 				$\ \ \ \ \ \ \ \ \#_{fail-obj-1} \gets \#_{fail-obj-1} + 1$
		 		
		 				$\ \ \ \ \ \ \ \ P_{obj-1} \gets max\{P_{obj-1-min}, $
		 				
		 				$\ \ \ \ \ \ \ \ \ \ \ \ \ \ \ \ \ \ \ \ \ \ \ \ P_{obj-1} - \#_{fail-obj-1} \cdot \Delta_{obj-1} \}$
		 				
		 		\EndIf
		 		\State \textbf{end if}
		 		
		 \EndIf		
		 \State \textbf{end if}

		 \If {$assigned \ to \ retrieve \ objects \ of \ type \ 2$}
		 		
		 		\If {$pick \ up \ success$}
		 		
		 				$\ \ \ \ \ \ \ \ \#_{succ-obj-2} \gets \#_{succ-obj-2} + 1$
		 				
		 				$\ \ \ \ \ \ \ \ \#_{fail-obj-2} \gets 0$
		 		
		 				$\ \ \ \ \ \ \ \ P_{obj-2} \gets min\{P_{obj-2-max}, $
		 				
		 				$\ \ \ \ \ \ \ \ \ \ \ \ \ \ \ \ \ \ \ \ \ \ \ \ P_{obj-2} + \#_{succ-obj-2} \cdot \Delta_{obj-2} \}$
		 				
		 		\EndIf
		 		\State \textbf{end if}
		 		
		 		\If {$pick \ up \ failure$}
		 					
		 				$\ \ \ \ \ \ \ \ \#_{succ-obj-2} \gets 0$
		 		
		 				$\ \ \ \ \ \ \ \ \#_{fail-obj-2} \gets \#_{fail-obj-2} + 1$
		 		
		 				$\ \ \ \ \ \ \ \ P_{obj-2} \gets max\{P_{obj-2-min}, $
		 				
		 				$\ \ \ \ \ \ \ \ \ \ \ \ \ \ \ \ \ \ \ \ \ \ \ \ P_{obj-2} - \#_{fail-obj-2} \cdot \Delta_{obj-2} \}$
		 				
		 		\EndIf
		 		\State \textbf{end if}
		 		
		 \EndIf		
		 \State \textbf{end if}
  		
  		 $P_1 \gets min\{P_{1-max}, P_1 + \#_{succ} \cdot \Delta \} $
  		 
	\EndIf		
	\State \textbf{end if}

 	\If {$leave \ nest \ failure$} 
 	
		 $\#_{fail} \gets \#_{fail} + 1$
		 
		 $\#_{succ} \gets 0$
		 
		 $P_1 \gets max\{P_{1-min}, P_1 - \#_{fail} \cdot \Delta \} $ 
		  
	 \EndIf	
	
	\State \textbf{end if}

	\end{algorithmic}
\end{algorithm}


\bigskip
\section{\textbf{Results}}

	The original algorithm is tested first, where robots treat each type of object equally. 
	
	At the beginning of the experiment, the value of $P_1$ for each robot is set to be a predefined certain value. The value of $P_1$ for each robot is updated through the experiment due to the performance of the robot. Whether task-allocation occurs or not can be observed in the distribution of $P_1$: if task allocation occurs, then at the end of the experiments some of the robots will have high $P_1$ while the others will have low $P_1$, and the distribution of $P_1$ will present two peaks; otherwise it will have only one peak.
	
	Then the modified algorithm is tested, where robots have the ability to distinguish the two types of object.
	
	Also, at the beginning of the second set of experiment, the values of $P_{obj-1}$ and $P_{obj-2}$ for each robot are also set to be predefined certain values. The values of $P_{obj-1}$ and $P_{obj-2}$ for each robot are also updated through the experiment due to the performance of the robot. Whether the sub-task-allocation occurs or not can be observed in the distribution of $P_{obj-1}$ and the distribution of $P_{obj-2}$: if sub-task-allocation occurs, then at the end of the experiments some of the foragers will have high $P_{obj-1}$ but low $P_{obj-2}$ while the other foragers will have high $P_{obj-2}$ but low $P_{obj-1}$, and there will be two peaks in both the distribution of $P_{obj-1}$ and the distribution of $P_{obj-2}$; otherwise there will be only one peak in each of the distributions.

\subsection{\textbf{Simulation Result of Using The Original Algorithm}}

\begin{figure}[H]
			\centering
			\includegraphics[scale=0.32]{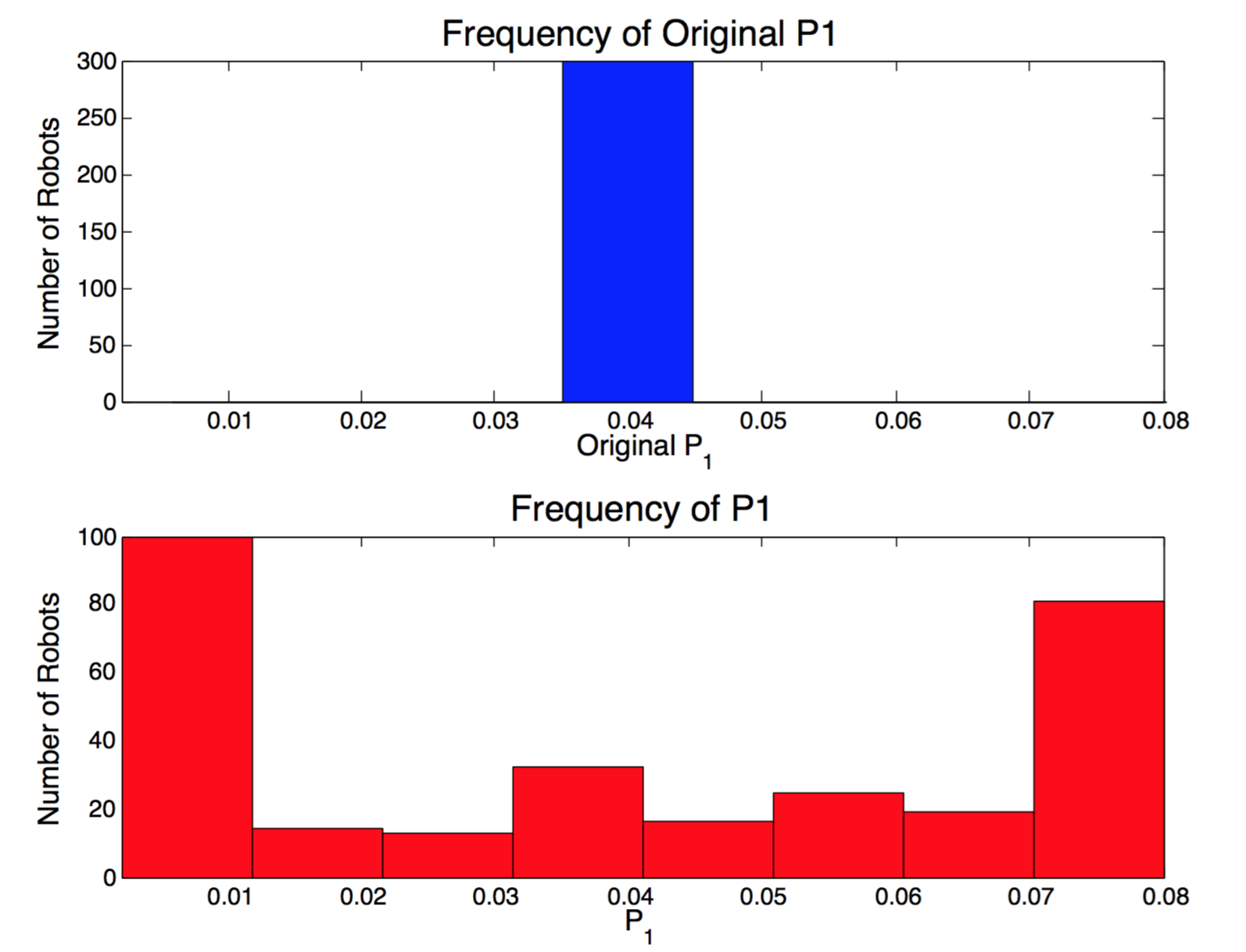}
			\caption{Frequency of $P_1$ Observed}
			\label{fig:result_1_1}	
		\end{figure}

	The frequencies of $P_1$ for the first set of experiment is presented in Figure \ref{fig:result_1_1}. The first sub-graph in Figure \ref{fig:result_1_1} shows the initial frequencies of $P_1$ at the beginning of the experiment, and all of the robots start from $P_1 = 0.04$. The second sub-graph in Figure \ref{fig:result_1_1} shows the retrieving result and its two-peak shape confirms that task-allocation has occurred.
	
\begin{figure}[H]
			\centering
			\includegraphics[scale=0.32]{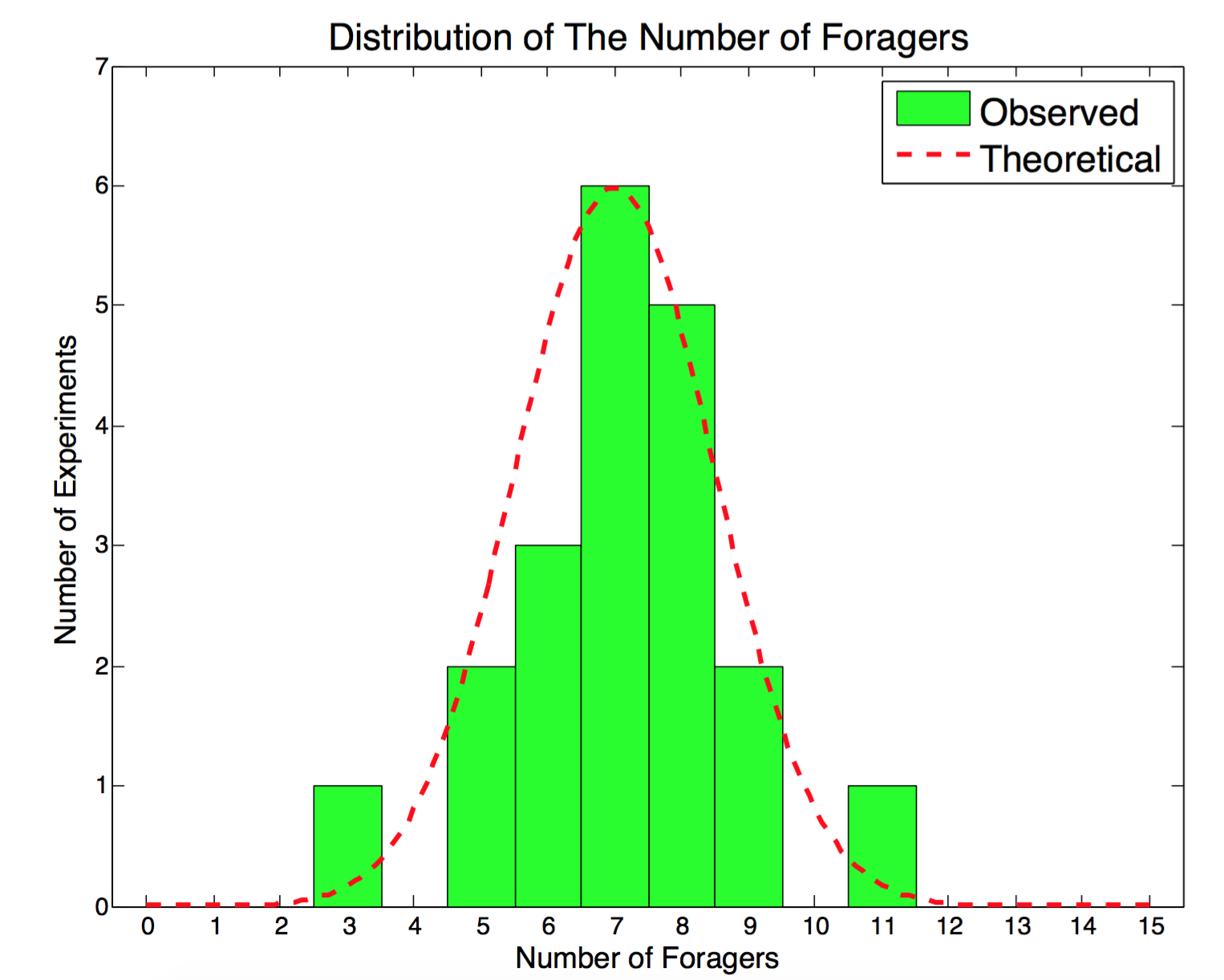}
			\caption{Distribution of The Number of Foragers Observed in Each Experiment Compared with The Theoretical Binomial Distribution}
			\label{fig:result_1_2}	
		\end{figure}

     The average value of the minimum $P_1$ and the maximum $P_1$ has been used as a point to separate robots into two groups, foragers and loafers. The distribution of the number of foragers observed in each experiment compared with the theoretical binomial distribution is presented in Figure \ref{fig:result_1_2}. Figure \ref{fig:result_1_2} shows that the profiles of the theoretical and the observed distributions are very similar and suggests that further experiments will confirm the matching.

\subsection{\textbf{Simulation Result of Using The Modified Algorithm}}

\begin{figure}[H]
			\centering
			\includegraphics[scale=0.32]{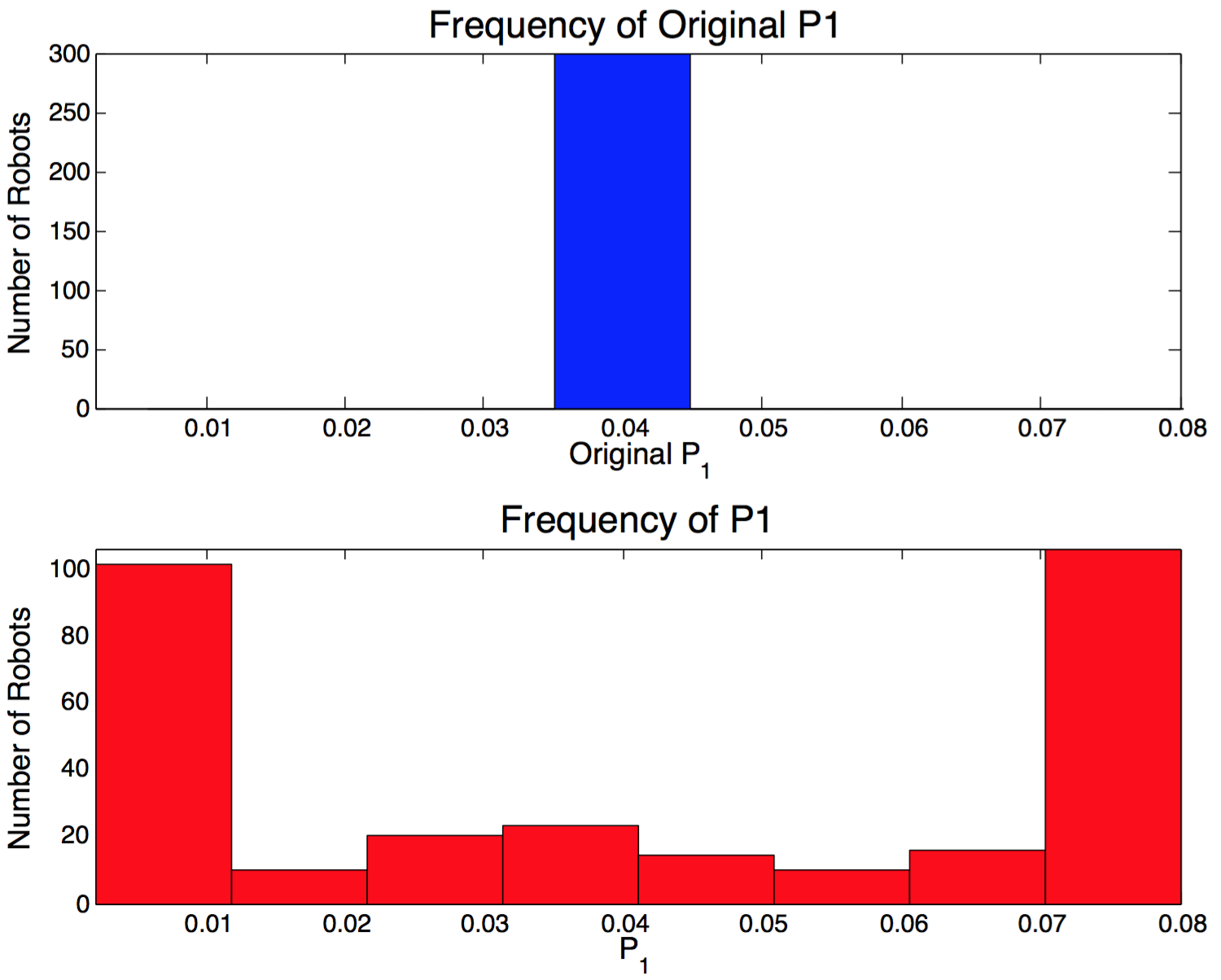}
			\caption{Frequency of $P_1$ Observed}
			\label{fig:result_2_1}	
		\end{figure}

\begin{figure}[H]
			\centering
			\includegraphics[scale=0.32]{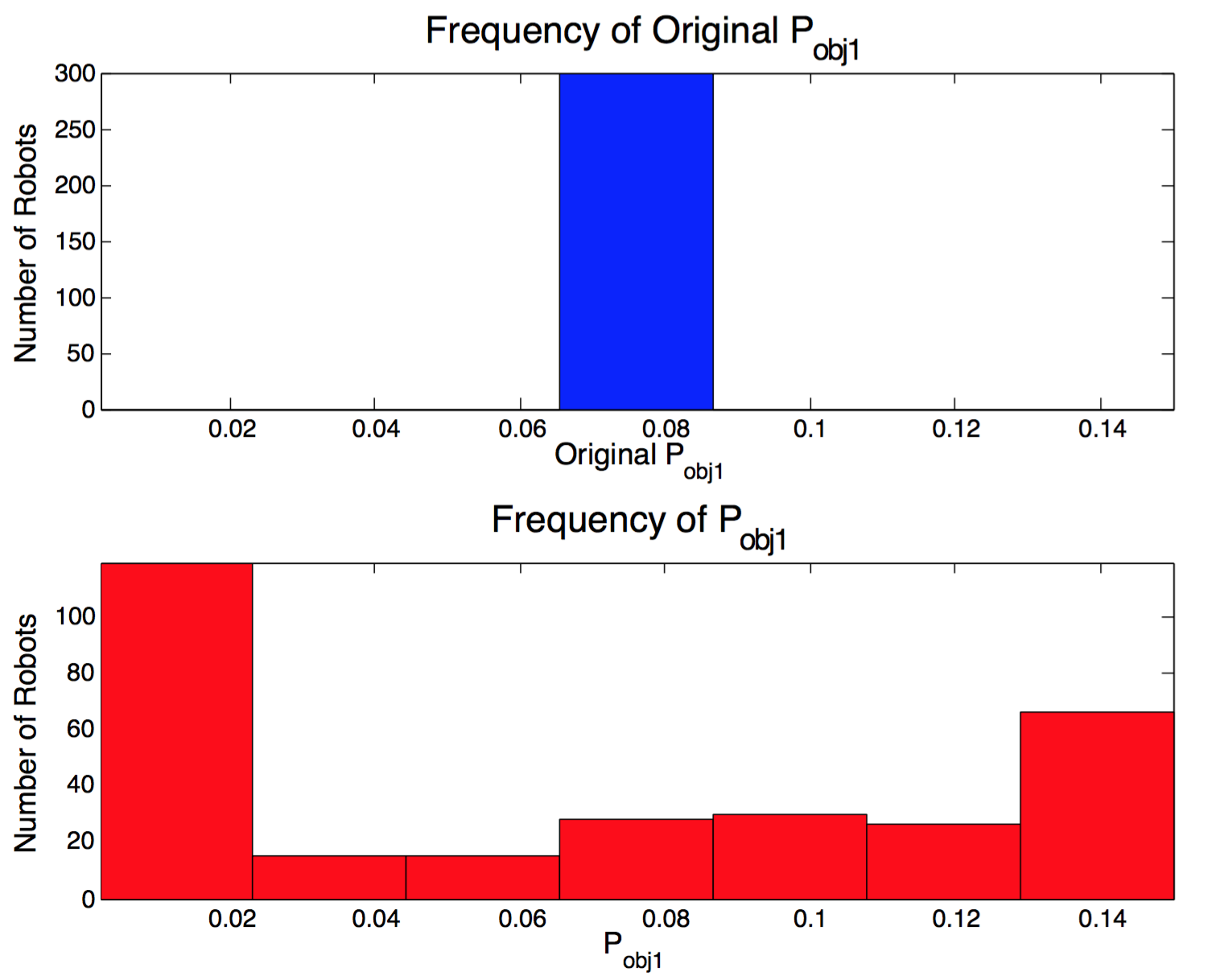}
			\caption{Frequency of $P_{obj-1}$ Observed}
			\label{fig:result_2_2}	
		\end{figure}

\begin{figure}[H]
			\centering
			\includegraphics[scale=0.32]{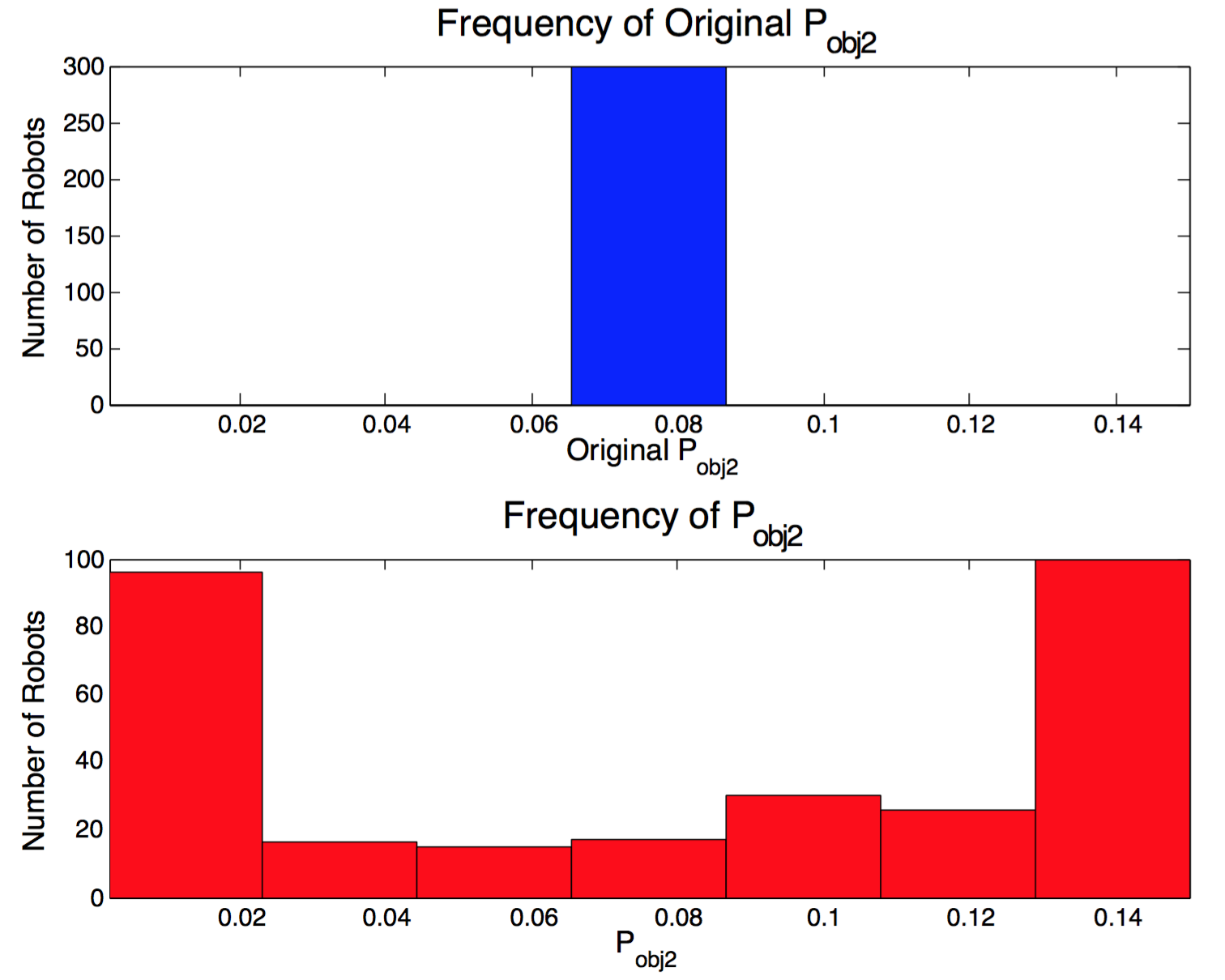}
			\caption{Frequency of $P_{obj-2}$ Observed}
			\label{fig:result_2_3}	
		\end{figure}

	The frequencies of $P_1$, $P_{obj-1}$ and $P_{obj-2}$ for the second set of experiment are presented separately in Figure \ref{fig:result_2_1}, Figure \ref{fig:result_2_2} and Figure \ref{fig:result_2_3}. 
	
	Figure \ref{fig:result_2_1} shows the "leaving-nest" result after 300s. The first sub-figure in Figure \ref{fig:result_2_1} shows the initial frequencies of $P_1$ at the beginning of the experiment. The two-peak shape in the second sub-figure in Figure \ref{fig:result_2_1}, which shows the updated $P_1$ after 300s, confirms that task allocation has occurred. 

	Figure \ref{fig:result_2_2} shows the "retrieving objects of type 1" result after 300s. The first sub-figure in Figure \ref{fig:result_2_2} shows the initial frequencies of $P_{obj-1}$ at the beginning of the experiment. The two-peak shape in the second sub-figure in Figure \ref{fig:result_2_2}, which shows the updated $P_{obj-1}$ after 300s, confirms that sub-task-allocation has occurred. 
	
	Figure \ref{fig:result_2_3} shows the "retrieving objects of type 2" result after 300s. The first sub-figure in Figure \ref{fig:result_2_3} shows the initial frequencies of $P_{obj-2}$ at the beginning of the experiment. The two-peak shape in the second sub-figure in Figure \ref{fig:result_2_3}, which shows the updated $P_{obj-2}$ after 300s, confirms that sub-task-allocation has occurred. 	
	
\begin{figure}[H]
			\centering
			\includegraphics[scale=0.32]{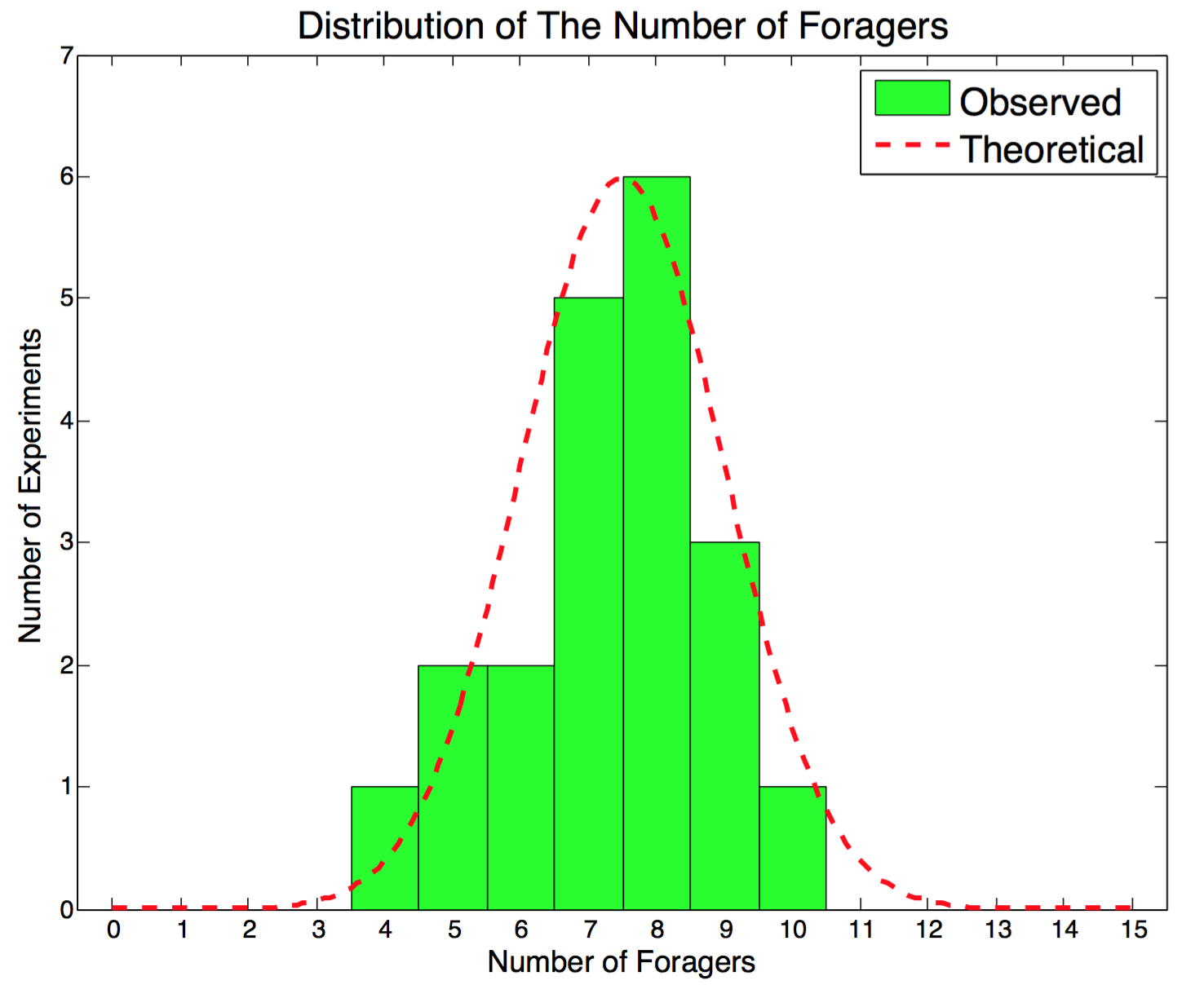}
			\caption{Distribution of The Number of Foragers Observed in Each Experiment Compared with The Theoretical Binomial Distribution}
			\label{fig:result_2_4}	
		\end{figure}

	Also the average value of the minimum $P_1$ and the maximum $P_1$ has been used as a point to separate robots into two groups, foragers and loafers. For the second set of experiment, the distribution of the number of foragers observed in each experiment compared with the theoretical binomial distribution is presented in Figure \ref{fig:result_2_4}. Figure \ref{fig:result_2_4} shows that the profiles of the theoretical and the observed distributions are very similar and suggests that further experiments will confirm the matching.

\begin{figure}
			\centering
			\includegraphics[scale=0.42]{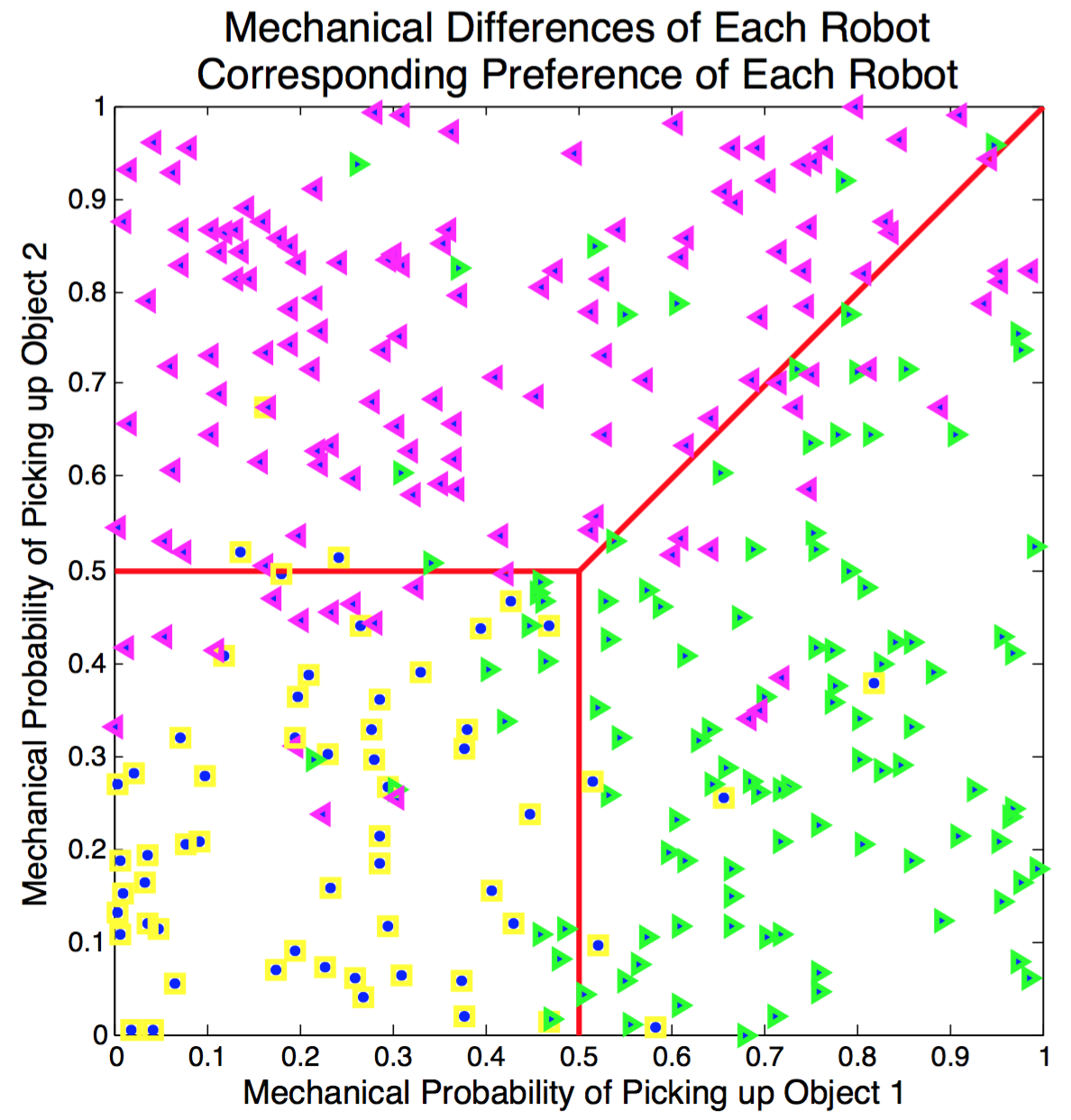}
			\caption{Robot Preference Corresponding to Mechanical Difference}
			\label{fig:result_2_5}	
		\end{figure}

	Robots have different mechanical capabilities of picking up each type of object. The capability is defined as a decimal number between 0 and 1, where 0 means not able to pick up the object and 1 means 100\% to pick up the object. Figure \ref{fig:result_2_5} presents the result of each robot's final preference corresponding to its mechanical capabilities. The x-axis represents each robot's capability of picking up objects of type 1 and the y-axis represents each robot's capability of picking up objects of type 2. Every point represents one robot, locating corresponding to its mechanical capabilities. There are three red lines in Figure \ref{fig:result_2_5}. If a robot has a mechanical capability of picking up objects of type 1 less than 0.5, it means the robot is supposed to be a "loafer of picking objects of type 1". Otherwise, the robot is supposed to be a "forager of picking objects of type 1". Likewise, if a robot has a mechanical capability of picking up objects of type 2 less than 0.5, it means the robot is supposed to be a "loafer of picking objects of type 2". Otherwise, the robot is supposed to be a "forager of picking objects of type 2". If a robot's mechanical capabilities of picking of both types of object less than 0.5, it means this robot is supposed to be a loafer of picking up both types of object. 
	
	Therefor, the robots located inside the bottom left square are supposed to be loafers, which have low $P_{obj-1}$ values and low $P_{obj-2}$ values.  The robots located inside the right trapezoid are supposed to be more likely to become foragers of picking up objects of type 1, which have high $P_{obj-1}$ values but low $P_{obj-2}$ values. The robots located inside the upper trapezoid are supposed to be more likely to become foragers of picking up objects of type 2, which have low $P_{obj-1}$ values but high $P_{obj-2}$ values. 
	
	After each experiment,  $P_{obj-1}$ and $P_{obj-2}$ of each robot are updated due to the performance of the corresponding robot. In each experiment, there are 15 robots, which means there are 15 $P_{obj-1}$ values and 15 $P_{obj-2}$ values. The average value of the minimum updated $P_{obj-1}$ and the maximum updated $P_{obj-1}$ is taken as a point to separated the robots ability of picking up object of type 1. Robots that have $P_{obj-1}$ values less than the average are considered to be loafers of picking up objects of type 1. Same to $P_{obj-2}$ values. If a robot's $P_{obj-1}$ and $P_{obj-2}$ are both smaller than the corresponding average values, this robot will be marked yellow, which indicates it is a loafer. Otherwise, $P_{obj-1}$ and $P_{obj-2}$ are compared. If a robot has $P_{obj-1}$ larger than $P_{obj-2}$, it will be marked green, which indicates this robot have a preference to pick up objects of type 1. If a robot has $P_{obj-1}$ larger than $P_{obj-2}$, it will be marked purple, which indicates this robot have a preference to pick up objects of type 2. Some robots may have both high $P_{obj-1}$ and high $P_{obj-2}$. However, there must be a larger value between the two values, so each robot would have a preference. 
	
	From Figure \ref{fig:result_2_5}, we can see that most of the robots that are supposed to be loafers (low mechanical capabilities of picking up both types of object) are marked yellow; most of the robots that are supposed to be foragers of picking up objects of type 1 (high mechanical capability of picking up objects of type 1 but low mechanical capability of picking up objects of type 2) are marked green; and most of the robots that are supposed to be foragers of picking up objects of type 2 (high mechanical capability of picking up objects of type 2 but low mechanical capability of picking up objects of type 2) are marked purple. This result proves that the sub-task-allocation happens and the implementation is correct.

\bigskip
\section{\textbf{Conclusion And Future Work}}

	The experiment proves that the task-allocation for leaving nest occurs and the sub-task-allocation for retrieving certain type of object also occurs.
	
	In the future, people can keep modifying the algorithm. For example, in the current modified algorithm, all objects are assumed to have same weight that is within the robot's holding capacity. In the future, objects can be assigned to have different weights, which may or may not overshoot the robot's holding capacity. Robots may work together to bring the heavy objects back to the central nest. For another example, in the current modified algorithm, all objects are located at different unchanged positions while the robots are searching. In the future, objects may move randomly in the experimental arena while the robot is searching.


\bigskip
\section*{\textbf{Acknowledgment}}

I cannot express enough thanks to my course advisor, Professor Michael Rubenstein, for their continued support and encouragement. I offer my sincere appreciation for the learning opportunities provided by Professor Rubenstein. My completion of this project could not have been accomplished without the support of Professor Rubenstein. My heartfelt thanks.



\begin{thebibliography}{1}

\bibitem{IEEEhowto:kopka}
Thomas H. Labella, Marco Dorigo and Jean-Louis Deneubourg. \emph{Self-Organised Task Allocation}.\hskip 1em plus
  0.5em minus 0.4em\relax Proceedings of the 7th International Symposium on Distributed Autonomous Robotic System, Toulouse, France, June, 2004.

\bibitem{IEEEhowto:kopka}
W. Agassounon and A. Martinoli. \emph{Efficiency and robustness of threshold- based distributed allocation algorithms in multi-agent systems.} \hskip 1em plus
  0.5em minus 0.4em\relax In C. Castelfranchi and W.L. Johnson, editors, Proceedings of the First International Joint Conference on Autonomous Agents and Multi-Agent Systems (AAMAS-02), pages 1090–1097. ACM Press, New York, NY, USA, 2002.

\bibitem{IEEEhowto:kopka}
T. Balch.  \emph{The impact of diversity on performance in multi-robot foraging.} \hskip 1em plus
  0.5em minus 0.4em\relax  In O. Etzioni, J.P. Mu ̈ller, and J.M. Bradshaw, editors, Proceedings of the Third International Conference on Autonomous Agents (Agents’99), pages 92–99. ACM Press, New York, NY, USA, 1999.

\bibitem{IEEEhowto:kopka}
S. Camazine, J.-L. Deneubourg, N.R. Franks, J. Sneyd, G. Theraulaz, and E. Bonabeau. \emph{Self-Organisation in Biological Systems.} \hskip 1em plus
  0.5em minus 0.4em\relax   Princeton University Press, Princeton, NJ, USA, 2001.


\end{thebibliography}
\end{document}